\documentclass{article} 
  \usepackage[T1]{fontenc}
  \usepackage{iclr2027_conference,times}
  \usepackage{amsmath,amssymb,amsfonts}
  \usepackage{graphicx}
  \usepackage{array}
  \usepackage{tabularx}
  \usepackage{booktabs}
  \usepackage{microtype}
  \usepackage{placeins}
  \usepackage{xurl}
  \usepackage{etoolbox}
  \usepackage[hidelinks]{hyperref}
  \hypersetup{pdftitle={Humanoid World Action Model With Joint State--Action Generation},pdfauthor={}}
  \AtBeginEnvironment{table}{\setlength{\belowcaptionskip}{5pt}}
  \AtBeginEnvironment{thebibliography}{\setlength{\bibsep}{6pt plus 2pt minus 1pt}}
\title{\raggedright HWAM: Humanoid World Action Model\\With Joint State--Action Generation}

\author{\textbf{Yan Yang\textsuperscript{1,*}, Jikun Rong\textsuperscript{1,*}, Minzhao Zhu\textsuperscript{1}, Zheyi Zhao\textsuperscript{1}, Qirui Hu\textsuperscript{1},} \\
  \textbf{Zihan Lan\textsuperscript{1}, Weixin Mao\textsuperscript{1}, Yinhao Li\textsuperscript{1}, Zhen Fu\textsuperscript{1, 2}, Hua Chen\textsuperscript{1,\ensuremath{\dagger}}} \\[10pt]
  \textbf{\textsuperscript{1}LimX Dynamics} \\
  \textbf{\textsuperscript{2}Southern University of Science and Technology (sustech.edu.cn)} \\[10pt]
  \textbf{\textsuperscript{*}Equal contribution. \textsuperscript{\ensuremath{\dagger}}Corresponding author.}
}

\iclrfinalcopy

\begin{document}

\maketitle

\begin{abstract}
Humanoid robots are a promising platform for general-purpose manipulation. Recent Vision-Language-Action (VLA) policies learn actions directly from multimodal observations, while World Action Models (WAMs) further incorporate future visual prediction to improve action generation. However, in hierarchical humanoid systems, VLA and WAM policies output reference actions that are subsequently realized through whole-body control, robot dynamics, balance, and contact. This hierarchy creates an action--execution gap: the reference produced by the policy can differ from the motion realized by the robot. Without explicitly modeling the realized body state, future visual prediction must jointly explain scene evolution and discrepancies between reference actions and executed motion, making it difficult to associate an action with its physical outcome.
We propose HWAM, a Humanoid World Action Model with joint state--action generation, which makes the robot's post-execution proprioceptive state an explicit prediction target. By jointly generating reference actions and their realized body states, HWAM directly incorporates supervision of executed motion into action learning. HWAM is trained through three complementary conditional paths. The Policy path jointly denoises state--action trajectories conditioned only on current observations, matching deployment conditions. Forward Dynamics Modeling (FDM) predicts future visual observations conditioned on actions and post-execution states, while Inverse Dynamics Modeling (IDM) reconstructs the joint trajectory from visual transitions. Together, these paths connect policy references, realized body motion, and visual outcomes.
HWAM achieves the highest success rate among evaluated baselines on three real-robot tasks on the LimX OLI humanoid. On Candy Picking, HWAM achieves a 70.6\% success rate, compared with 43.3\% for Fast-WAM. Code, dataset, and project page are available at \url{https://hwam.vercel.app/}.
\end{abstract}

\section{Introduction}

General-purpose humanoid robots are expected to understand human instructions and execute diverse tasks in complex physical environments.
Recent Vision-language-action (VLA) models have become a promising approach to this problem by unifying visual perception, language understanding, and action generation in a single multimodal policy. Systems such as PaLM-E, RT-2, and OpenVLA show that semantic knowledge from large vision-language models can support robot control and transfer across tasks and scenes \citep{palm-e, rt2, openvla}. World Action Models (WAMs) further introduce future visual or latent-state prediction into action generation, allowing predicted futures to provide additional supervision or information for control \citep{wam_survey, uwm, wam_zero}. These advances have moved robot policies beyond pure action regression toward predictive and generative control.

\begin{figure}[!htbp]
  \centering
  \includegraphics[width=\textwidth]{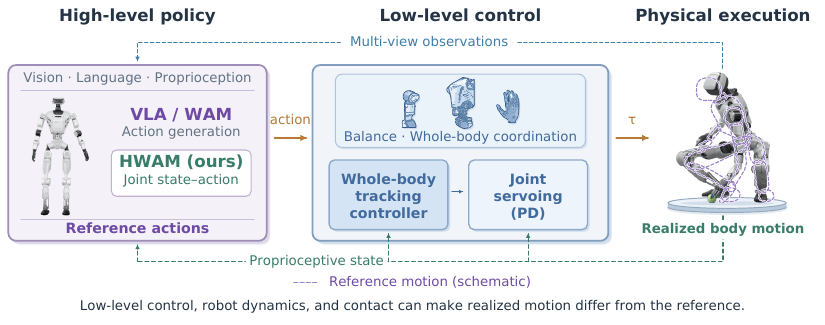}
  \caption{\textbf{Hierarchical control interface in HWAM.} The high-level VLA/WAM/HWAM policy produces reference actions, which are adapted by low-level whole-body tracking and joint servoing before physical execution. The measured post-execution proprioceptive state provides feedback to the high-level policy.}
  \label{fig:hierarchy}
\end{figure}

General-purpose humanoid manipulation requires both task-level motion decisions and execution-level whole-body dynamics control. We adopt a hierarchical control architecture: a high-level vision--language--action policy generates reference trajectories in kinematic space from visual observations, language instructions, and the current proprioceptive state, while a low-level whole-body tracking controller uses real-time proprioceptive feedback to adapt these references to execution dynamics and drives the robot through joint servoing. This division allows the high-level policy to learn task behavior from demonstrations that primarily provide poses and motion trajectories, while reusing the low-level controller's whole-body coordination and feedback control capabilities \citep{humanplus, omnih2o}. Accordingly, the action \(a_t\) in this paper denotes a motion reference supplied to the low-level tracking controller, rather than a torque applied directly to an actuator, and it is not assumed to be realizable point by point. Figure~\ref{fig:hierarchy} summarizes this hierarchical control interface.

This hierarchical interface also introduces a discrepancy between the high-level reference and the actual body response. The low-level tracking policy adapts reference motion to the current body state, contact conditions, and dynamic constraints, so the realized motion can deviate from the reference. For example, a reference trajectory may request simultaneous trunk motion and arm extension; adjustments to trunk motion made by the low-level controller to maintain balance can also change the hand trajectory at contact. Thus, even a kinematically feasible reference does not ensure the intended task outcome after low-level execution. The high-level policy needs to learn the body responses produced by low-level tracking and adaptation to robot dynamics, so that reference generation can account for actual execution outcomes. We refer to the discrepancy between a high-level reference action and its physical realization as the \emph{action--execution gap}.

This gap complicates action learning through future visual prediction.  A
policy reference can differ from the state realized after low-level
execution.  If this intermediate state is not modeled, the model must learn
simultaneously how the policy action is physically realized and how the
realized motion changes the scene.  This makes the relation between the policy
action and its future visual outcome difficult to learn.  HWAM addresses this problem by explicitly placing
the post-execution state in the action-generation target and then connecting
the joint state--action trajectory to visual transitions in both directions.

Dataset statistics reveal larger discrepancies between commanded joint references and measured proprioceptive states in mobile humanoid manipulation than in stationary bimanual manipulation, including under steady-state conditions (Section~\ref{sec:execution_gap}).
This observation motivates representing the post-execution proprioceptive
state explicitly instead of asking a policy to relate a high-level action
directly to its downstream visual consequence.

Prior work has explored body-state representations and prediction for
humanoid control~\citep{humanoid_policy, hex}. We propose HWAM, a
humanoid-oriented World Action Model that incorporates
post-execution states into its trajectory generation target. Joint generation
is a natural extension of the WAM action stream: state and action share a
temporal structure and can be modeled by the same trajectory generator. At
each temporal position, we align action \(a_{t+h}\) with proprioceptive state
\(s_{t+h+1}\) and represent them as the joint state--action target
\begin{equation}
  \mathbf{Y}_t[h]=[s_{t+h+1};a_{t+h}].
  \label{eq:intro_joint_target}
\end{equation}
The state and action channels are concatenated before token projection and
jointly denoised by a shared state--action generator. This gives action
generation access to the evolving state estimate throughout denoising. We
study how the resulting joint target interacts with visual training
objectives to improve closed-loop task performance.

Building on conditional world-action modeling~\citep{uwm, cosmos3}, HWAM
incorporates post-execution states into three training paths. The Policy path
jointly denoises state--action trajectories conditioned on current visual
observations, language, and proprioception, without generating future video.
This supervises policy references together with their measured execution
outcomes. Forward Dynamics Modeling
(FDM) supplements this objective by conditioning future visual prediction on
actions and post-execution states, explicitly accounting for realized body
motion when predicting scene changes. Inverse Dynamics Modeling (IDM) adds
supervision in the reverse direction, using visual transitions to jointly
predict actions and the body states they produce. The three paths connect
policy references to both execution states and visual outcomes. At deployment,
predicted states continue to participate in joint denoising, while only the
actions are sent to the low-level controller.

Our evaluation asks whether the action--execution gap is observable in robot
data, whether a joint state--action target becomes effective when combined
with bidirectional visual grounding, and whether the resulting policy improves
real-robot manipulation. We evaluate stationary and mobile manipulation on the
LimX OLI humanoid and a task-specific clothes-folding setting on ALOHA~\citep{aloha}. Our main
contributions are:

\begin{enumerate}
    \item \textbf{Post-execution state--action generation.} We pair each
    high-level action with its temporally aligned post-execution
    proprioceptive state and denoise the two channels as one joint target.

    \item \textbf{Execution-state grounding.} Post-execution states serve as
    conditions for future-video prediction in FDM and as joint prediction
    targets with actions in IDM and Policy, connecting execution-state
    supervision to visual transitions.

    \item \textbf{Controlled real-robot evidence.} A factorial ablation
    identifies the interaction between the joint target and the three-path
    training formulation, and real-robot experiments evaluate the resulting
    model on humanoid and bimanual manipulation.
\end{enumerate}
\section{Related Work}

\subsection{Vision-Language-Action Models}

Vision-language-action (VLA) models unify visual perception, language understanding, and robot control. PaLM-E integrates embodied observations into a multimodal language model for reasoning and planning \citep{palm-e}. RT-2 transfers vision-language knowledge to robot control via action tokenization, while OpenVLA explores open-source VLA learning from large-scale demonstrations \citep{rt2, openvla}. Recent systems improve generalization through heterogeneous data and continuous action generation. For instance, $\pi_{0.5}$ achieves open-world generalization via multi-source co-training, enabling manipulation in novel environments \citep{pi05}, and its successor $\pi_{0.7}$ introduces a steerable generalist foundation model with emergent compositional generalization and cross-embodiment transfer \citep{pi07}. In NVIDIA's GR00T series, GR00T N1 is an open foundation model for generalist humanoid reasoning and skills, combining a pretrained vision-language module with a diffusion Transformer action generator for cross-embodiment manipulation \citep{groot_n1}. Its updates, GR00T N1.5 and N1.7, enhance instruction following and upgrade the VLM backbone to Cosmos-Reason2-2B (Qwen3-VL) with large-scale human and simulation pretraining, improving generalization and long-horizon reasoning \citep{groot_n15, groot_n17}.

\subsection{World Action Models and Joint Generative Policies}

World Action Models incorporate future prediction into policy learning.
Existing approaches include cascaded models that first generate future
observations and then infer actions, as well as joint models that generate
future visual variables and actions within a shared generative framework
\citep{wam_survey}. Unified World Models (UWM) couple video and action
diffusion and use modality-specific diffusion steps to represent policies,
forward dynamics, inverse dynamics, and video generation
\citep{uwm}. Cosmos 3 distinguishes forward dynamics, inverse dynamics, and
a policy mode that jointly predicts video and action~\citep{cosmos3}.
Motus integrates understanding, video, and action experts in a
Mixture-of-Transformers architecture and supports multiple conditional
generation modes within one model \citep{motus}. Fast-WAM retains video
co-training while predicting actions from current observations without
explicit future-video synthesis at inference time \citep{fast_wam}.

Joint generative trajectory modeling provides a related precedent. Diffuser
models state--action trajectories with a diffusion model and generates complete
trajectories through iterative denoising \citep{diffuser}. Cosmos Policy
jointly predicts actions and a future state containing proprioception and
images~\citep{cosmos_policy}.

\subsection{Body-State Representations for Humanoid Control}

Body-state representations help humanoid policies coordinate high-dimensional
configurations, multiple effectors, and contact-dependent motion. HumanPlus
and OmniH2O use hierarchical control to connect task-level motion
references with whole-body execution \citep{humanplus, omnih2o}.
\citet{humanoid_policy} use a shared state--action representation
to learn humanoid policies from human and robot demonstrations.
HEX combines humanoid-aligned body-part representations with a Unified
Proprioceptive Predictor and fuses predictive body-state features with
visual-language information for whole-body control \citep{hex}. These works
establish body state as a useful representation for transferring motion
knowledge and learning coordinated behavior.


\section{Method}

\subsection{Hierarchical Control Interface and Overview}

\begin{figure}[!htbp]
  \centering
  \includegraphics[width=\textwidth]{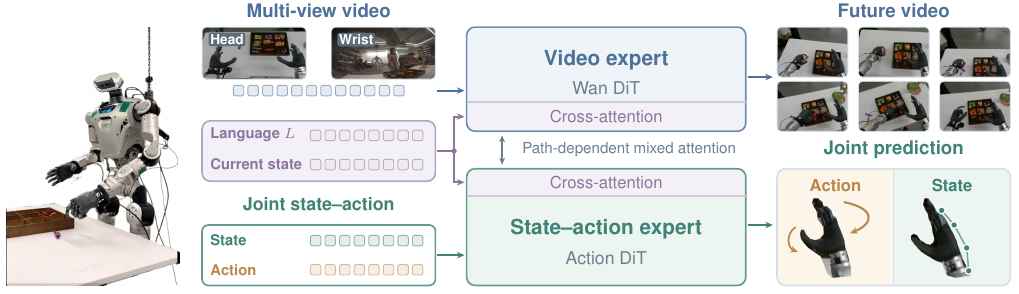}
  \caption{\textbf{Overview of HWAM.} The high-level action $a_t$ is executed through
  the low-level whole-body controller, producing the measured post-execution
  state $s_{t+1}$. HWAM represents the joint state--action pair as a single
  token. A frozen causal video VAE provides current and future visual latents,
  while the video and state--action experts interact through path-dependent
  mixed attention.}
  \label{fig:architecture}
\end{figure}

We adopt a hierarchical interface in which a high-level VLA/WAM policy
outputs kinematic reference actions \(a_t\) (rather than direct actuator
torques), and a low-level whole-body tracking controller adapts them using
proprioceptive feedback. We write \(s_{t+1}\) for the proprioceptive state
measured after execution. HWAM treats each reference action and its aligned post-execution state as a
joint prediction target, generated by a shared state--action generator. A video expert grounds this joint trajectory in
visual transitions through path-dependent mixed attention. Training uses three
conditional paths: FDM predicts future video from a demonstrated joint
state--action trajectory, IDM predicts the joint trajectory from visual
evidence, and Policy generates only the state--action trajectory conditioned
on current observations, without generating future video. Figure~\ref{fig:architecture} gives an overview.
\subsection{Joint Post-Execution State--Action Chunk Target}

Under the hierarchical interface above, the high-level action $a_t$ is a
motion reference, while $s_{t+1}$ is the proprioceptive state measured after
low-level execution. Task success therefore depends on the state actually
reached after execution; an action that is valid in kinematic space is not
sufficient.

At time $t$, the robot receives synchronized multi-view observations
$O_t=\{o_t^{(v)}\}_{v=1}^{V}$, a language instruction $L$, and its measured
proprioceptive state $s_t$. For training targets, $t$ indexes synchronized
demonstration samples: $a_t$ denotes the high-level motion reference sent to
the low-level tracking controller, and $s_{t+1}$ is the proprioceptive state
recorded at the next sample. This naturally couples the action $a_t$ with its
execution outcome $s_{t+1}$.

For a prediction horizon of $H$ steps, we define the action and state
chunks as
\begin{equation}
  \mathbf A_t=[a_t,\ldots,a_{t+H-1}],
  \qquad
  \mathbf S_t=[s_{t+1},\ldots,s_{t+H}].
  \label{eq:chunks}
\end{equation}
We adopt a simple and effective formulation for joint prediction: the model
learns to predict a joint state--action chunk by concatenating the state and
action targets along the feature dimension. Let $\mathbf Y_t$ be the joint
target:
\begin{equation}
  \begin{aligned}
    \mathbf Y_t
      &=\operatorname{Concat}(\mathbf S_t,\mathbf A_t)
        \in\mathbb R^{H\times(d_s+d_a)},\\
    \mathbf Y_t[h]
      &=[s_{t+h+1};a_{t+h}],
        \qquad h=0,\ldots,H-1.
  \end{aligned}
  \label{eq:joint_target}
\end{equation}

HWAM places the post-execution state in the same generation target as the
action, extending the WAM trajectory stream with state channels. The two are
concatenated before token projection and modeled through a shared trajectory
representation. Thus, at every denoising step, the action channels
can access the current partially denoised state channels, while state
supervision can influence the representation used for action generation.

This construction provides post-execution state supervision through a shared
state--action representation. It does not explicitly learn a low-level
dynamics model or impose a separate physical-consistency constraint. Moreover,
the joint target is not assumed to be useful in isolation: the next subsection
describes how it is grounded in visual transitions. 

\subsection{Architecture}

  HWAM consists of two interacting experts: a video expert that models visual
  transitions and a state--action expert that generates
  $\mathbf Y_t$. Given a synchronized multi-view clip $\mathbf O_t^V$, a frozen
  causal Wan VAE~\citep{wan} produces the current and future visual latents
  $\mathbf Z_t^0$ and $\mathbf Z_t^+$, denoted collectively by
  $\mathbf Z_t=[\mathbf Z_t^0;\mathbf Z_t^+]$. The language instruction $L$ and
  the current proprioceptive state $s_t$ are encoded as the shared conditioning
  context $\mathbf C_t$. Both experts receive $\mathbf C_t$, while their access
  to $\mathbf Z_t^0$, $\mathbf Z_t^+$, and $\mathbf Y_t$ is determined by the
  selected training path.

  The state--action expert extends the action stream used in
  action-only WAMs. It operates on the target
  $\mathbf Y_t=\operatorname{Concat}(\mathbf S_t,\mathbf A_t)$ defined
  above, where each temporal token contains the post-execution state and its
  corresponding action. The state and action channels are projected
  into a shared trajectory representation and jointly denoised with the same
  state--action time $\tau_{SA}$. Consequently, the partially denoised state
  channels can influence action generation throughout denoising within the
  shared trajectory stream.

  The video and state--action experts use separate Transformer
  parameters and exchange information through path-dependent masked
  cross-attention, following the mixed-expert interface of prior WAMs
  \citep{fast_wam}. The masks determine the visibility of $\mathbf Z_t^0$,
  $\mathbf Z_t^+$, and $\mathbf Y_t$ in the FDM, IDM, and Policy paths,
  allowing the same architecture to support trajectory-to-video,
  vision-to-trajectory, and deployment-matched policy learning. The detailed
  visibility patterns and objectives are described in the next subsection.

\subsection{Bidirectional State--Action--Vision Grounding}

\begin{figure}[!htbp]
  \centering
  \includegraphics[width=\textwidth]{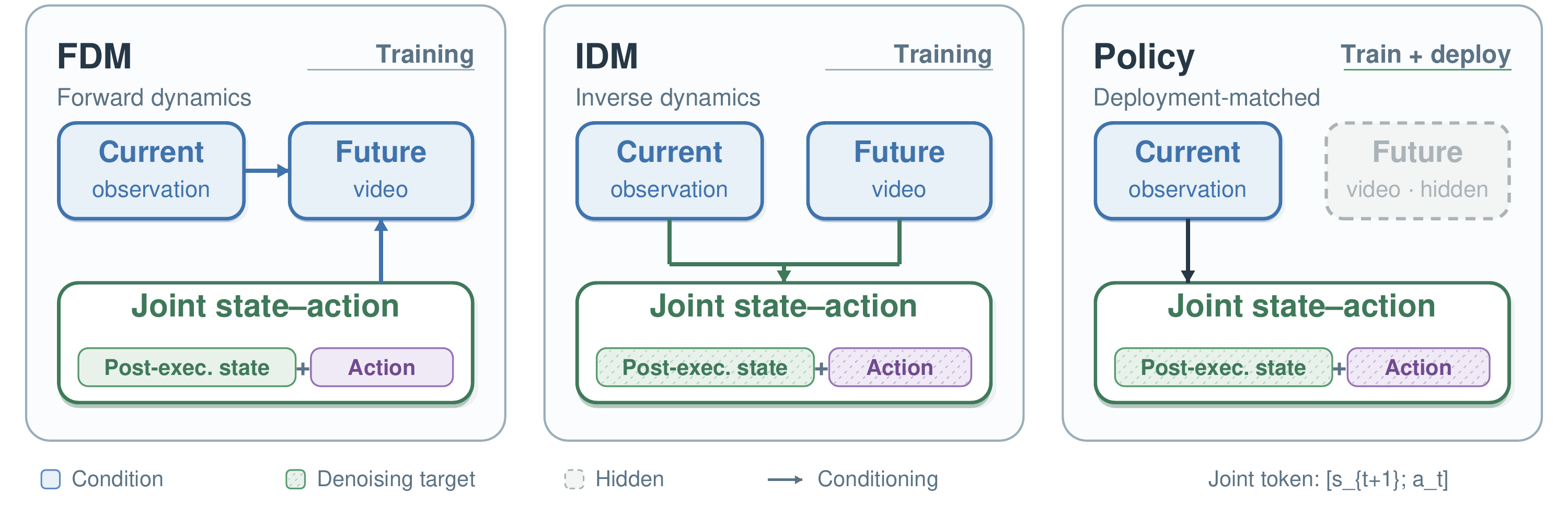}
  \caption{\textbf{Path-dependent corruption and attention.} FDM predicts future visual observations from the joint state–action trajectory, whereas IDM reconstructs the joint trajectory from visual transitions, providing complementary bidirectional grounding between visual dynamics and state–action trajectories. Policy matches deployment by conditioning only on current observations.}
  \label{fig:paths}
\end{figure}

The joint state--action target is trained under three conditional paths
that share the same experts and parameters but differ in generation targets
and visible conditions. Let \(\mathcal V_0=\mathbf Z_t^0\),
\(\mathcal V_+=\mathbf Z_t^+\), and \(\mathcal J=\mathbf Y_t\) denote
current-video tokens, future-video tokens, and joint state--action tokens.
Figure~\ref{fig:paths} summarizes the three paths and their attention masks.

\paragraph{Training roles.}
The Policy path conditions only on current observations and generates the
joint state--action trajectory without future video, matching deployment. FDM
predicts future video from actions and realized post-execution states, so
visual prediction accounts for physical execution rather than the command
alone. IDM reverses this direction by predicting actions and post-execution
states from visual transitions.

Together, these paths make the post-execution state both an input for
explaining visual outcomes and a supervised output coupled to action
generation. The factorial ablation in Section~\ref{sec:ablation} evaluates
this combined training design against VGM/Policy, which combines video
generation modeling (VGM) with Policy training.

\paragraph{Conditional flow-matching objectives.}
We use conditional flow matching \citep{flow_matching} for both generated
modalities. For a clean target $\mathbf X$, Gaussian noise
$\boldsymbol\epsilon_X\sim\mathcal N(0,\mathbf I)$, and noise level
$\tau\in[0,1]$, the corrupted target and target velocity are
\begin{equation}
  \mathbf X_\tau=(1-\tau)\mathbf X+\tau\boldsymbol\epsilon_X,
  \qquad
  \mathbf U_X=\boldsymbol\epsilon_X-\mathbf X.
  \label{eq:flow_interpolation}
\end{equation}
We apply this construction to the future-video target $\mathbf Z_t^+$ and
the joint trajectory target $\mathbf Y_t$, with independent noise levels
$\tau_V$ and $\tau_{SA}$ and corresponding velocities $\mathbf U_V$ and
$\mathbf U_{SA}$. State and action channels share $\tau_{SA}$. We use
elementwise mean squared error, denoted by $\operatorname{MSE}$, for each
generated target. State and action errors are averaged together without
additional channel or timestep weights.

The clean contextual conditions for each path are
\(\mathbf Q_F=(\mathbf Z_t^0,\mathbf Y_t,\mathbf C_t)\),
\(\mathbf Q_I=(\mathbf Z_t^0,\mathbf Z_t^+,\mathbf C_t)\), and
\(\mathbf Q_P=(\mathbf Z_t^0,\mathbf C_t)\).
The video and state--action velocity predictors, denoted by $v_\theta^V$
and $v_\theta^{SA}$, process the noisy target conditioned on both its noise
level and the path-specific context $\mathbf Q_m$. Their losses are
\begin{equation}
  \begin{aligned}
    \mathcal L_{\mathrm F}
      &=\mathbb E\!\left[
        \operatorname{MSE}\!\left(v_\theta^V(\mathbf Z_{t,\tau_V}^{+}\mid\tau_V,\mathbf Q_F),
          \mathbf U_V\right)\right],\\
    \mathcal L_{\mathrm I}
      &=\mathbb E\!\left[
        \operatorname{MSE}\!\left(v_\theta^{SA}(\mathbf Y_{t,\tau_{SA}}\mid\tau_{SA},\mathbf Q_I),
          \mathbf U_{SA}\right)\right],\\
    \mathcal L_{\mathrm P}
      &=\mathbb E\!\left[
        \operatorname{MSE}\!\left(v_\theta^{SA}(\mathbf Y_{t,\tau_{SA}}\mid\tau_{SA},\mathbf Q_P),
          \mathbf U_{SA}\right)\right].
  \end{aligned}
  \label{eq:path_losses}
\end{equation}
The expectations are over demonstration samples, Gaussian noise, and the
noise levels sampled for the generated modality. FDM corrupts and supervises
only future-video targets; IDM and Policy corrupt and supervise the joint
trajectory. Conditioning variables remain clean and are excluded from the
loss. In Policy, future-video tokens are hidden and only state--action tokens
are denoised, during both training and deployment.

We sample one path $m\in\{\mathrm F,\mathrm I,\mathrm P\}$ per minibatch,
with $\pi_m=1/3$ and $\lambda_m=1$, and optimize
\begin{equation}
  \mathcal L_{\mathrm{HWAM}}
    =\mathbb E_{m\sim\pi}
      \left[\lambda_m\mathcal L_m\right]
  \quad
  =\tfrac13(\mathcal L_{\mathrm F}+\mathcal L_{\mathrm I}+\mathcal L_{\mathrm P}).
  \label{eq:total}
\end{equation}
\section{Experiments}
\label{sec:experiments}

We evaluate HWAM on three real-robot tasks with the LimX OLI humanoid and on
offline open-loop diagnostics. We compare with vision--language--action (VLA)
and world-model-based policies, and conduct a controlled ablation on the
mobile manipulation task to examine the joint state--action target with
FDM/IDM/Policy training. ALOHA results are reported in
Appendix~\ref{app:aloha}.

\subsection{Action--Execution Gap in Robot Data}
\label{sec:execution_gap}

Before evaluating policy performance, we quantify how high-level actions are
realized in recorded robot trajectories. We compare each commanded joint
reference \(a_t\) with its aligned post-execution proprioceptive state
\(s_{t+1}\) on the LimX OLI mobile humanoid and the stationary ALOHA bimanual
platform. Both settings show non-zero action--state errors, including residual
tracking errors under approximately constant joint references
(LimX OLI: \(4.420^\circ\) direct, \(2.885^\circ\) steady-state; ALOHA:
\(0.648^\circ\) direct, \(0.402^\circ\) steady-state). These data motivate
modeling the post-execution state rather than assuming exact tracking. The
full measurement protocol, alignment rules, and per-joint statistics are
provided in Appendix~\ref{app:execution_gap}.
\subsection{Experimental Setup}
\label{sec:experimental_setup}
\label{sec:oli_protocol}

\noindent\textbf{Platform, tasks, and data.}
We use the LimX OLI, a full-size humanoid with a 31-DoF whole-body action
space (including two 7-DoF arms) and BrainCo dexterous hands. The OLI suite
contains two stationary manipulation tasks, \emph{Candy Picking} and
\emph{Object Collection}, and one mobile whole-body manipulation task,
\emph{Plush-Toy Picking}. 
placement, repeated reaching, and mobile whole-body coordination. The
task-specific training sets contain 1,783, 4,220, and 404 episodes,
respectively. Candy Picking has an additional held-out validation set of 422
episodes for offline diagnostics; these episodes are excluded from training.
Further task definitions, control demands, and completion criteria are given
in Appendix~\ref{app:oli_tasks}.
\begin{figure}[t]
  \centering
  \includegraphics[width=\linewidth]{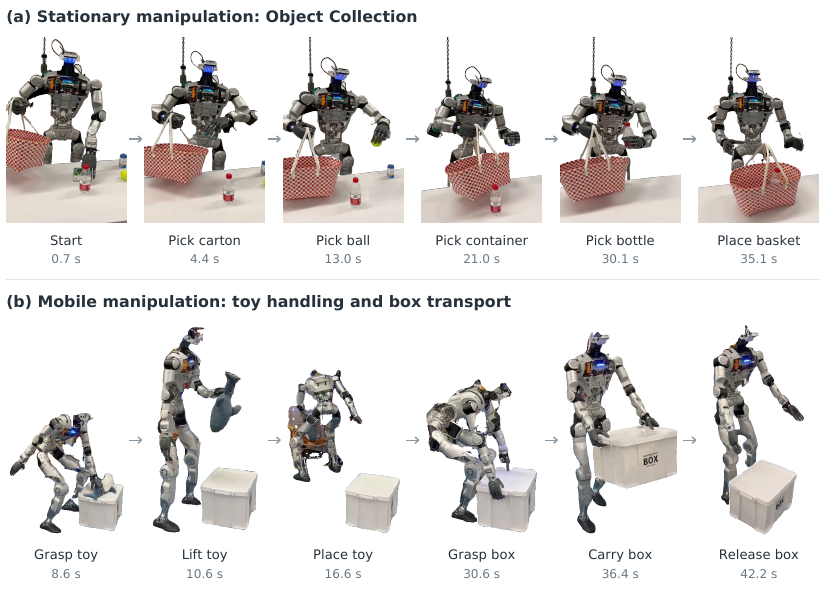}
  \caption{\textbf{Examples from the OLI evaluation suite. }The stationary manipulation
  example shows repeated tabletop object collection, while the mobile manipulation example
  combines object handling with changes in body posture and position. Each
  row follows a separate video timeline; backgrounds are simplified and
  panels are cropped for visualization.}
  \label{fig:oli_task_sequences}
\end{figure}

\noindent\textbf{Baselines.}
We compare HWAM with five policies on OLI: GR00T N1.5~\citep{groot_n15},
$\pi_{0.5}$~\citep{pi05}, Cosmos3-Edge~\citep{cosmos3},
DiT4DiT~\citep{dit4dit}, and Fast-WAM~\citep{fast_wam}. Fast-WAM is the
closest prior WAM baseline because it also uses video co-training while
generating actions without future-video synthesis at deployment. The ALOHA
appendix additionally reports results for $\pi_0$~\citep{pi0} and
SmolVLA~\citep{smolvla}.

\noindent\textbf{Evaluation protocol.}
On OLI, we evaluate policies through physical robot trials and report success
rate as $100c/n$, where $c$ is the number of successful trials and $n$ is
the total number of trials for a method and task. Completion criteria, the
Candy Picking color-to-region mapping, trial counts, and confidence intervals
are summarized in Appendix~\ref{app:oli_tasks}
and Appendix~\ref{app:oli_counts}.

\noindent\textbf{Implementation.}
Our implementation builds on the FluxVLA Engine codebase~\citep{fluxvla_engine}.
HWAM uses a frozen Wan2.2 video VAE~\citep{wan, wan22}, a WanVideoDiT video
expert, an ActionDiT state--action expert, and UMT5-XXL language embeddings
~\citep{umt5}. For the OLI experiments, both experts are fine-tuned for six
epochs with AdamW at a learning rate of $10^{-4}$, using a global batch size
of 128 on 16 A800 GPUs. FDM, IDM, and Policy are sampled uniformly. ALOHA-specific settings are provided in
Appendix~\ref{app:aloha}.

\subsection{Real-Robot Evaluation}
\label{sec:main_results}

Table~\ref{tab:oli_main} summarizes the OLI real-robot results. The
robot-data column records whether the compared model was pretrained with
additional robot data as reported for the corresponding baseline.

\begin{table}[!htbp]
    \centering
    \small
    \setlength{\tabcolsep}{3pt}
    \caption{Real-robot evaluation results (\%) on the three OLI tasks.
    An em dash denotes an unreported or unverified task result.}
    \label{tab:oli_main}
    \begin{tabular}{@{}lcccc@{}}
      \toprule
      Method & Robot-data pretraining
        & Candy Picking & Object Collection & Plush-Toy Picking \\
      \midrule
      Cosmos3-Edge & Yes & 15.8 & 0.0 & 56.7 \\
      GR00T N1.5 & Yes & 40.5 & 42.5 & 60.0 \\
      $\pi_{0.5}$ & Yes & 62.5 & 45.0 & 60.7 \\
      Fast-WAM & No & 43.3 & 0.0 & 60.0 \\
      DiT4DiT & No & 52.5 & 15.0 & \textemdash \\
      \textbf{HWAM (Ours)} & No
        & \textbf{70.6} & \textbf{46.7} & \textbf{73.3} \\
      \bottomrule
    \end{tabular}
  \end{table}
HWAM obtains the highest observed success rate on all three OLI tasks.
Relative to $\pi_{0.5}$, the observed improvements are 8.1 percentage points
on Candy Picking, 1.7 points on Object Collection, and 12.6 points on
Plush-Toy Picking. These results show the effectiveness of the complete
HWAM policy across the evaluated stationary and mobile manipulation tasks.
On Object Collection, Cosmos3-Edge and Fast-WAM fail to establish the basket grasp
required by the task, accounting for their 0\% full-task success rates.

\subsection{Open-Loop Prediction Diagnostics}
\label{sec:open_loop}

\begin{figure}[!htbp]
  \centering
  \includegraphics[width=1.0\textwidth]{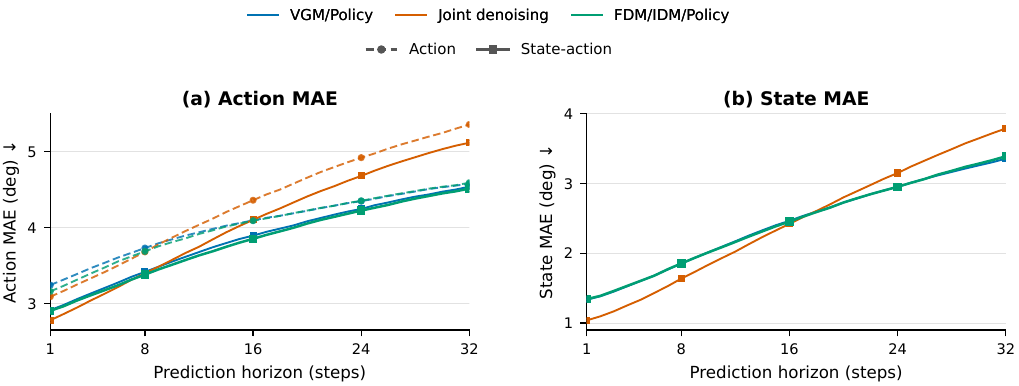}
  \caption{\textbf{Open-loop action (a) and state (b) MAE on Candy Picking
  over a 32-step horizon.} Errors are averaged over 31 shared joints on the
  held-out episodes and reported in degrees. Dashed and solid curves denote
  action-only and joint state--action prediction, respectively. Lower is
  better.}
  \label{fig:open_loop_mae}
\end{figure}

We evaluate open-loop action and state prediction on the held-out Candy
Picking episodes described in Section~\ref{sec:experimental_setup}. We use
four denoising steps and compute MAE in degrees over the first 31 shared
joints. Figure~\ref{fig:open_loop_mae} shows MAE at each position of the
32-step chunk.
Averaged over the 32 chunk positions, HWAM achieves the lowest action MAE
($3.816^\circ$), compared with $4.024^\circ$ for action-only
FDM/IDM/Policy. Adding state supervision reduces action MAE across all
evaluated training recipes. HWAM and the VGM/Policy state--action variant
show similar state MAE.

For future-video prediction, we use windows immediately before grasp
closure in 12 held-out episodes. Under predicted trajectory conditions,
PSNR ranges from 20.345 to 20.844 and HWAM shows no video-quality advantage.
Under ground-truth conditions, conditioning on both actions and
post-execution states improves PSNR from 20.559 to 23.336 over the
action-only model, supporting the value of realized-state conditioning.
Full results are in Appendix~\ref{app:open_loop_video}.
\subsection{Controlled Ablation on Plush-Toy Picking}
\label{sec:ablation}

We evaluate four matched configurations on Plush-Toy Picking, the mobile
whole-body manipulation task in the OLI suite. This task couples object
handling with body movement and balance control, providing a whole-body
setting to examine the interaction between the joint state--action target and
the training recipe. The configurations vary two choices: the prediction
target is either action-only or joint state--action, and the training recipe
is either VGM/Policy or FDM/IDM/Policy. All other settings are fixed.
VGM/Policy jointly optimizes future-video generation and the deployment-matched
Policy objective, whereas FDM/IDM/Policy samples separate paths for forward
dynamics, inverse dynamics, and policy learning.

\begin{table}[!htbp]
  \centering
  \small
  \setlength{\tabcolsep}{5pt}
  \caption{Controlled ablation on Plush-Toy Picking.}
  \label{tab:plush_ablation}
  \begin{tabular*}{\linewidth}{@{\extracolsep{\fill}}lllc@{}}
    \toprule
    Configuration & Prediction target & Training recipe & Success (\%) \\
    \midrule
    Action + VGM/Policy
      & Action & VGM/Policy & 60.0 \\
    State--Action + VGM/Policy
      & Joint state--action & VGM/Policy & 45.5 \\
    Action + FDM/IDM/Policy
      & Action & FDM/IDM/Policy & 55.0 \\
    \textbf{HWAM}
      & Joint state--action & FDM/IDM/Policy & \textbf{73.3} \\
    \bottomrule
  \end{tabular*}
\end{table}

In this ablation, the effect of the joint target depends on the visual training recipe. Under
VGM/Policy, adding state--action prediction decreases success from 60.0\% to
45.5\%. Under FDM/IDM/Policy, the same change increases success from
55.0\% to 73.3\%. The observed difference between these two effects is
32.8 percentage points. This target-by-recipe interaction supports combining
joint state--action generation with FDM/IDM/Policy training in the evaluated
setting. The complete HWAM configuration achieves the highest success rate
among the four configurations.
The ablation evaluates the combined formulation in this task and does not
isolate the contribution of individual paths or architectural choices.

\section{Conclusion}

We introduced HWAM, which jointly generates high-level actions and their
temporally aligned post-execution proprioceptive states. The method is
motivated by a measurable action--execution gap: low-level control, dynamics,
and contact can cause realized body motion to differ from policy references.
The joint target is a simple extension of the WAM trajectory stream, trained
with FDM and IDM to connect execution states and action references to visual
transitions, alongside a Policy path that generates only state--action
trajectories conditioned on current observations.
HWAM achieves the highest observed success rate on the three evaluated OLI
tasks. The ablation on the mobile manipulation task shows that adding the
joint target improves task success under FDM/IDM/Policy training and reduces it under VGM/Policy,
supporting the combined design in this setting. Additional ALOHA results
demonstrate its applicability to bimanual clothes folding. HWAM does not
explicitly identify low-level dynamics, and its
predicted states are not sent to the controller; they serve as jointly
generated variables that connect high-level actions to their physical and
visual consequences.

\section*{AI Use Statement}
Generative AI tools were used to assist with code implementation, manuscript
drafting and editing, and pre-submission feedback on the research methodology
and experimental reporting, including interpretation of reported results and
identification of relevant literature. The authors take responsibility for
the final content of this work, including the implementation, experimental
results, conclusions, and any AI-assisted text or other artifacts.

\bibliographystyle{iclr2027_conference}
\bibliography{references}

\clearpage
\appendix

\section{Action--Execution Gap Analysis}
\label{app:execution_gap}

We compare each commanded joint reference \(a_t\) with its aligned
post-execution proprioceptive state \(s_{t+1}\) on the LimX OLI mobile
humanoid and the stationary ALOHA bimanual platform.
Table~\ref{tab:execution_gap_summary} summarizes the gap. Both settings show
non-zero action--state errors, including residual tracking errors when the
commanded joint references are approximately constant. These data motivate
modeling the post-execution state rather than assuming exact tracking.

\begin{table}[!htbp]
  \centering
  \small
  \caption{Action--execution gap measured by mean absolute error (MAE) in
  degrees. Direct MAE compares the joint reference \(a_t\) with the measured
  post-execution joint state \(s_{t+1}\). Steady-state MAE uses the same pairing
  under approximately constant joint references.}
  \label{tab:execution_gap_summary}
  \begin{tabular}{lcc}
    \toprule
    Platform & Direct MAE & Steady-state MAE \\
    \midrule
    LimX OLI (mobile humanoid) & 4.420 & 2.885 \\
    ALOHA (stationary bimanual) & 0.648 & 0.402 \\
    \bottomrule
  \end{tabular}
\end{table}

\section{Task Suite and Data}
\label{app:oli_tasks}

Table~\ref{tab:oli_tasks} summarizes the three OLI tasks used in the main
evaluation. The episode counts refer to the robot demonstrations used for
training each task.

\begin{table}[!htbp]
  \centering
  \small
  \renewcommand{\arraystretch}{1.3} 
  \setlength{\tabcolsep}{6pt}
  \caption{OLI task suite, training data, and completion criteria.}
  \label{tab:oli_tasks}
  \begin{tabularx}{\linewidth}{@{}>{\raggedright\arraybackslash}p{0.17\linewidth}
                  >{\raggedright\arraybackslash}X
                  >{\centering\arraybackslash}p{0.15\linewidth}
                  >{\raggedright\arraybackslash}X@{}}
    \toprule
    Task & Control demands & \shortstack{Training\\episodes} & Completion criterion \\
    \midrule
    Candy Picking
      & Precise grasping and spatially targeted placement
      & 1,783
      & Pick the instructed candy and place it in its assigned tray region. \\
    \addlinespace[3pt]
    Object Collection
      & Repeated reaching and grasping across objects
      & 4,220
      & Place all four tabletop objects in the basket. \\
    \addlinespace[3pt]
    Plush-Toy Picking
      & Mobile object handling with whole-body coordination
      & 404
      & Complete the specified plush-toy handling sequence. \\
    \bottomrule
  \end{tabularx}
\end{table}
\paragraph{Training data and observations.}
\label{app:oli_data}
For Candy Picking, the training set contains 1,783 demonstrations covering
the three instructed candy colors. A separate set of 422 held-out episodes
is used for offline validation and is excluded from training. For Object
Collection and Plush-Toy Picking, the training sets contain 4,220 and 404
episodes, respectively. The OLI policy receives synchronized
RGB observations from the head and wrist cameras, a language instruction, and
the current proprioceptive state. Task success rates are measured in physical
robot trials; the held-out demonstrations are used for prediction diagnostics.

\FloatBarrier
\section{Additional OLI Results}
\label{app:oli_additional}
\subsection{Future-Video Prediction Diagnostics}
\label{app:open_loop_video}

We evaluate deployment-matched future-video prediction on 12 held-out
episodes from the same Candy Picking task. The evaluation window is
immediately before grasp closure. All conditions in
Table~\ref{tab:open_loop_video} are predicted rather than taken from the
ground-truth future trajectory. Here, \emph{joint denoising} denotes joint
video--trajectory denoising: the baseline corrupts and denoises both streams
within one training mode, without separating the directional FDM and IDM paths.

\begin{table}[!htbp]
  \centering
  \small
  \caption{End-to-end future-video prediction without ground-truth future
  state or action conditions. Higher is better.}
  \label{tab:open_loop_video}
  \begin{tabular}{lcc}
    \toprule
    Configuration & PSNR & SSIM \\
    \midrule
    VGM + Action & 20.690 & 0.7420 \\
    VGM + State--Action & 20.673 & 0.7409 \\
    Joint denoising + Action & 20.844 & 0.7453 \\
    Joint denoising + State--Action & 20.670 & 0.7440 \\
    FDM/IDM/Policy + Action & 20.345 & 0.7361 \\
    HWAM & 20.501 & 0.7370 \\
    \bottomrule
  \end{tabular}
\end{table}

HWAM does not obtain a video-quality advantage in this end-to-end comparison.
PSNR ranges from 20.345 to 20.844 across configurations, with HWAM at 20.501.
These measurements characterize future-video prediction and do not explain
the differences in closed-loop task performance.

Table~\ref{tab:open_loop_video_oracle} compares the action-only
FDM/IDM/Policy model and HWAM under ground-truth trajectory conditions.

\begin{table}[!htbp]
  \centering
  \small
  \caption{Future-video prediction with ground-truth trajectory conditions
  on held-out Candy Picking episodes. Higher is better.}
  \label{tab:open_loop_video_oracle}
  \begin{tabular}{llc}
    \toprule
    Model & Condition & PSNR \\
    \midrule
    Action-only & Ground-truth actions & 20.559 \\
    HWAM & Ground-truth states and actions & 23.336 \\
    \bottomrule
  \end{tabular}
\end{table}

HWAM achieves a PSNR of 23.336, compared with 20.559 for the action-only
model. This result supports the value of realized-state conditioning for
predicting visual outcomes, consistent with the action--execution gap between
policy references and physical motion.
\subsection{Trial Counts and Uncertainty}
\label{app:oli_counts}

Table~\ref{tab:oli_candy_counts} supplements the Candy Picking result with
available trial counts and 95\% Wilson confidence intervals. These intervals
assume independent Bernoulli trials and do not measure variation across
training seeds.

\begin{table}[!htbp]
  \centering
  \small
  \caption{Candy Picking trial counts and 95\% Wilson confidence intervals
  for methods with available counts.}
  \label{tab:oli_candy_counts}
  \begin{tabular}{lccc}
    \toprule
    Method & Successes / trials & SR (\%) & 95\% CI (\%) \\
    \midrule
    GR00T N1.5 & 49/121 & 40.5 & [32.2, 49.4] \\
    $\pi_{0.5}$ & 75/120 & 62.5 & [53.6, 70.6] \\
    Fast-WAM & 52/120 & 43.3 & [34.8, 52.3] \\
    HWAM & 84/119 & 70.6 & [61.9, 78.0] \\
    \bottomrule
  \end{tabular}
\end{table}

\subsection{Motion and Workspace Breakdowns}
\label{app:oli_spatial}

Table~\ref{tab:oli_spatial} reports available Candy Picking success rates by
motion direction and workspace region. These are within-task breakdowns, not
evaluations on a held-out spatial split.

\begin{table}[!htbp]
  \centering
  \small
  \caption{Candy Picking success rates (\%) by (a) motion direction and
  (b) workspace region. CH/CV: cross-horizontal/vertical; X-P/X-S:
  primary/secondary diagonal.}
  \label{tab:oli_spatial}
  \setlength{\tabcolsep}{3pt}

  \begin{minipage}[t]{0.48\textwidth}
    \centering
    \begin{tabular}{lrrrr}
      \toprule
      \multicolumn{5}{c}{(a) Motion directions} \\
      \midrule
      Method & CH & CV & X-P & X-S \\
      \midrule
      GR00T N1.5 & 50.0 & 25.8 & 40.0 & 46.7 \\
      $\pi_{0.5}$ & 60.0 & 66.7 & 80.0 & 43.3 \\
      Fast-WAM & 33.3 & 43.3 & 46.7 & 50.0 \\
      HWAM & 82.8 & 63.3 & 76.7 & 60.0 \\
      \bottomrule
    \end{tabular}
  \end{minipage}\hfill
  \begin{minipage}[t]{0.48\textwidth}
    \centering
    \begin{tabular}{lrrrrr}
      \toprule
      \multicolumn{6}{c}{(b) Workspace regions} \\
      \midrule
      Method & L1 & L2 & C & B2 & B3 \\
      \midrule
      GR00T N1.5 & 40.0 & 58.3 & 41.7 & 45.8 & 16.7 \\
      $\pi_{0.5}$ & -- & -- & -- & -- & -- \\
      Fast-WAM & 70.8 & 83.3 & 41.7 & 8.3 & 12.5 \\
      HWAM & 83.3 & 95.8 & 66.7 & 65.2 & 41.7 \\
      \bottomrule
    \end{tabular}
  \end{minipage}
\end{table}
\FloatBarrier
\section{Evaluation on ALOHA Clothes Folding}
\label{app:aloha}

\begin{figure}[!htbp]
  \centering
  \includegraphics[width=\linewidth]{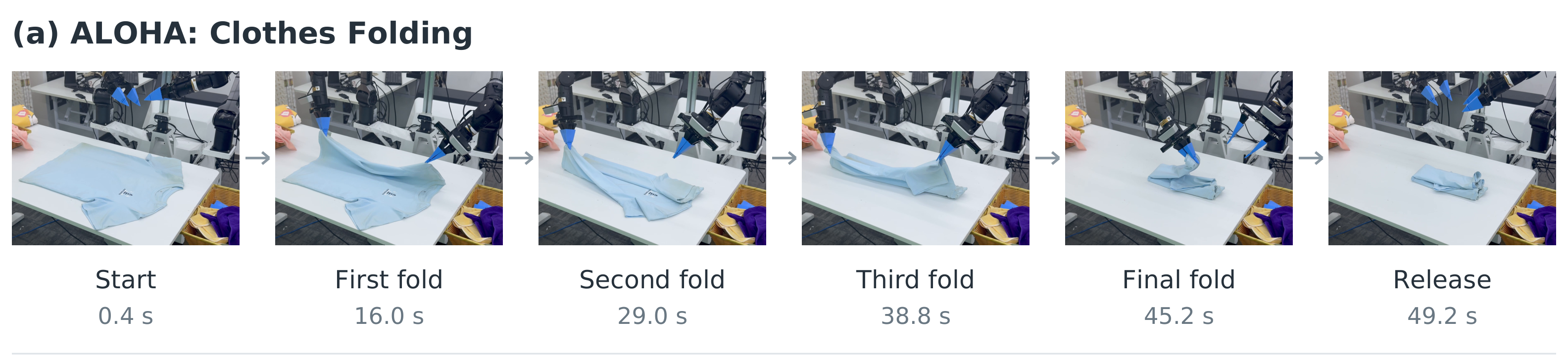}
  \caption{ALOHA Clothes Folding task. The sequence shows the robot
  performing successive bimanual folds on a spread garment, from the initial
  grasp (Start, 0.4\,s) through three intermediate folds to the final fold
  and release (49.2\,s).}
  \label{fig:aloha_folding_steps}
\end{figure}

\paragraph{Task and data.}
We evaluate HWAM on the ALOHA bimanual platform~\citep{aloha} using
\emph{Clothes Folding}, which requires coordinated grasping and successive
folds of a spread garment. The task-specific training set contains 640
demonstrations. Observations comprise RGB images from a high-mounted camera
and two wrist cameras, the proprioceptive state, and a language instruction.
This is a task-specific evaluation on ALOHA rather than zero-shot transfer
from the OLI platform. Figure~\ref{fig:aloha_folding_steps} illustrates the
task.

\paragraph{Training and evaluation.}
The ALOHA configuration retains joint state--action prediction and uniform
FDM/IDM/Policy sampling. The video and state--action experts are
fine-tuned for six epochs with AdamW at a learning rate of $10^{-4}$ and a
per-device batch size of eight, while the video VAE remains frozen.
Clothes Folding is evaluated using a task-completion score on a 0--100 scale,
with partial credit for completed folding stages.

\begin{table}[!htbp]
  \centering
  \small
  \setlength{\tabcolsep}{5pt}
  \caption{ALOHA Clothes Folding task-completion scores on a 0--100 scale.
  Scores include partial credit for completed folding stages. Higher is better.}
  \label{tab:aloha_results}
  \begin{tabular}{lc}
    \toprule
    Method & Clothes Folding score \\
    \midrule
    GR00T N1.5 & 78.75 \\
    GR00T N1.7 & 58.75 \\
    $\pi_0$ & 65.00 \\
    $\pi_{0.5}$ & 77.00 \\
    Fast-WAM & 66.25 \\
    SmolVLA & 51.25 \\
    \textbf{HWAM (Ours)} & \textbf{86.00} \\
    \bottomrule
  \end{tabular}
\end{table}

HWAM achieves the highest reported Clothes Folding task-completion score
among the evaluated methods. This result supports the applicability of joint
state--action generation to bimanual deformable-object manipulation
beyond the humanoid platform, while the task-specific training means that it
should not be interpreted as zero-shot cross-platform transfer.

\end{document}